# Learning to Bluff

Evan Hurwitz and Tshilidzi Marwala

*Abstract*— The act of bluffing confounds game designers to this day. The very nature of bluffing is even open for debate, adding further complication to the process of creating intelligent virtual players that can bluff, and hence play, realistically. Through the use of intelligent, learning agents, and carefully designed agent outlooks, an agent can in fact learn to predict its opponents' reactions based not only on its own cards, but on the actions of those around it. With this wider scope of understanding, an agent can in learn to bluff its opponents, with the action representing not an "illogical" action, as bluffing is often viewed, but rather as an act of maximising returns through an effective statistical optimisation. By using a TD(λ) learning algorithm to continuously adapt neural network agent intelligence, agents have been shown to be able to learn to bluff without outside prompting, and even to learn to call each other's bluffs in free, competitive play.

## I. INTRODUCTION

WHILE many card games involve an element of bluffing, simulating and fully understanding bluffing yet remains one of the most elusive tasks presented to the game design engineer. The entire process of bluffing relies on performing a task that is unexpected, and is thus misinterpreted by one's opponents. For this reason, static rules are doomed to failure since once they become predictable, they cannot be misinterpreted. In order to create an artificially intelligent agent that can bluff, one must first create an agent that is capable of learning. The agent must be able to learn not only about the inherent nature of the game it is playing, but also must be capable of learning trends emerging from its opponent's behaviour, since bluffing is only plausible when one can anticipate the opponent's reactions to one's own actions.

Firstly the game to be modelled will be detailed, with the reasoning for its choice being explained. The paper will then detail the system and agent architecture, which is of paramount importance since this not only ensures that the correct information is available to the agent, but also has a direct impact on the efficiency of the learning algorithms utilised. Once the system is fully illustrated, the actual learning of the agents is shown, with the appropriate findings detailed.

E. Hurwitz is with the School of Electrical and Information Engineering at the University of Witwatersrand, Johannesburg, South Africa; e-mail: e.hurwitz@ee.wits.ac.za).

Prof. T. Marwala is Carl Emily Fuchs Chair of Systems and Control at the School of Electrical and Information Engineering at the University of Witwatersrand, Johannesburg, South Africa; e-mail: t.marwala@ee.wits.ac.za ).

## II. LERPA

The card game being modelled is the game of Lerpa [4]. While not a well-known game, its rules suit the purposes of this research exceptionally well, making it an ideal testbed application for intelligent agent Multi-Agent Modelling (MAM). The rules of the game first need to be elaborated upon, in order to grasp the implications of the results obtained. Thus, the rules for Lerpa now follow.

The game of *Lerpa* is played with a standard deck of cards, with the exception that all of the 8s, 9s and 10s are removed from the deck. The cards are valued from greatest- to least-valued from ace down to 2, with the exception that the 7 is valued higher than a king, but lower than an ace, making it the second most valuable card in a suit. At the end of dealing the hand, the dealer has the choice of *dealing himself in* – which entails flipping his last card over, unseen up until this point, which then declares which suit is the *trump suit* [4]. Should he elect not to do this, he then flips the next card in the deck to determine the trump suit. Regardless, once trumps are determined, the players then take it in turns, going clockwise from the dealer's left, to elect whether or not to play the hand (to *knock*), or to drop out of the hand, referred to as *folding* (if the Dealer has *dealt himself in*, as described above, he is then automatically required to play the hand). Once all players have chosen, the players that have elected to play then play the hand, with the player to the dealer's left playing the first card. Once this card has been played, players must then play *in suit* – in other words, if a heart is played, they must play a heart if they have one. If they have none of the required suit, they may play a trump, which will win the trick unless another player plays a higher trump. The highest card played will win the trick (with all trumps valued higher than any other card) and the winner of the trick will lead the first card in the next trick. At any point in a hand, if a player has the Ace of trumps and can legally play it, he is then required to do so [4]. The true risk in the game comes from the betting, which occurs as follows:

At the beginning of the round, the dealer pays the table 3 of whatever the basic betting denomination is (referred to usually as 'chips'). At the end of the hand, the chips are divided up proportionately between the winners, i.e. if you win two tricks, you will receive two thirds of whatever is in the pot. However, if you stayed in, but did not win any tricks, you are said to have been *Lerpa'd*, and are then required to match whatever was in the pot for the next hand, effectively costing you the pot. It is in the evaluation of this risk that most of the true skill in *Lerpa* lies.

## III. LERPA MAM

As with any optimisation system, very careful consideration needs to be taken with regards to how the system is structured, since the implications of these decisions can often result in unintentional assumptions made by the system created. With this in mind, the Lerpa Multi-Agent System (MAS) has been designed to allow the maximum amount of freedom to the system, and the agents within, while also allowing for generalisation and swift convergence in order to allow the intelligent agents to interact unimpeded by human assumptions, intended or otherwise.

### A. System Overview

The game is, for this model, going to be played by four players. Each of these players will interact with each other indirectly, by interacting directly with the *table*, which is their shared environment, as depicted in Fig. 1.

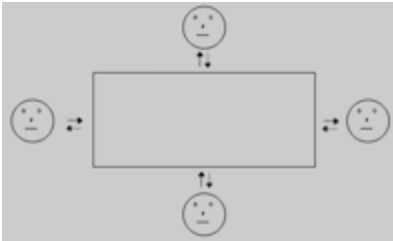

Fig. 1. System interactions

Over the course of a single hand, an agent will be required to make three decisions, once at each interactive stage of the game. These three decision-making stages are:
1) To play the hand, or drop (*knock* or *fold*).
2) Which card to play first.
3) Which card to play second.

Since there is no decision to be made at the final card, the hand can be said to be effectively finished from the agent's perspective after it has played its second card (or indeed after the first decision should the agent fold). Following on the TD($\lambda$) algorithm [5], each agent will update its own neural network at each stage, using its own predictions as a reward function, only receiving a true reward after its final decision has been made. This decision making process is illustrated below, in Fig. 2.

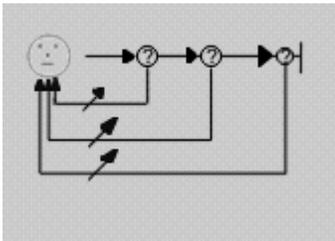

Fig. 2. Agent learning scheme

With each agent implemented as described, the agents can now interact with each other through their shared environment, and will continuously learn upon each interaction and its consequent result.

Each hand played will be viewed as an independent, stochastic event, and as such only information about the current hand will be available to the agent, who will have to draw on its own learned knowledge base to draw deductions not from previous hands

### B. Agent AI Design

A number of decisions need to be made in order to implement the agent artificial intelligence (AI) effectively and efficiently. The type of learning to be implemented needs to be chosen, as well as the neural network architecture [7]. Special attention needs to be paid to the design of the inputs to the neural network, as these determine what the agent can 'see' at any given point. This will also determine what assumptions, if any, are implicitly made by the agent, and hence cannot be taken lightly. Lastly, this will determine the dimensionality of the network, which directly affects the learning rate of the network, and hence must obviously be minimized.

*1) Input Parameter Design:* In order to design the input stage of the agent's neural network, one must first determine all that the network may need to know at any given decision-making stage. All inputs, in order to optimise stability, are structured as binary-encoded inputs. When making its first decision, the agent needs to know its own cards, which agents have stayed in or folded, and which agents are still to decide [9]. It is necessary for the agent to be able to determine which specific agents have taken their specific actions, as this will allow for an agent to learn a particular opponent's characteristics, something impossible to do if it can only see a number of players in or out. Similarly, the agent's own cards must be specified fully, allowing the agent to draw its own conclusions about each card's relative value. It is also necessary to tell the agent which suit has been designated the trumps suit, but a more elegant method has been found to handle that information, as will be seen shortly. Fig. 3 below illustrates the initial information required by the network.

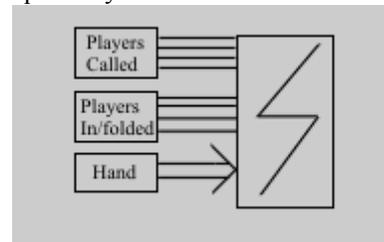

Fig. 3. Basic input structure

The agent's hand needs to be explicitly described, and the obvious solution is to encode the cards exactly, i.e. four suits, and ten numbers in each suit, giving forty possibilities for each card. A quick glimpse at the number of options available shows that a raw encoding style provides a sizeable problem of dimensionality, since an encoded hand can be one of $40^3$ possible hands (in actuality, only $^{40}P_3$ hands could be selected, since cards cannot be repeated, but the raw encoding scheme would in fact allow for repeated cards, and hence $40^3$ options would be available). The first thing to notice is that only a single deck of cards is being used, hence no card can ever be repeated in a hand. Acting on this principle, consistent ordering of the hand means that the base dimensionality of the hand is greatly reduced, since it is now combinations of cards that are represented, instead of permutations. The number of combinations now represented is $^{40}C_3$. This seemingly small change from $^nP_r$ to $^nC_r$ reduces the dimensionality of the representation by a factor of r!, which in this case is a factor of 6. Furthermore, the representation of cards as belonging to discrete suits is not optimal either, since the game places no particular value on any suit by its own virtue, but rather by virtue of which suit is the trump suit. For this reason, an alternate encoding scheme has been determined, rating the 'suits' based upon the makeup of the agent's hand, rather than four arbitrary suits. The suits are encoded as belonging to one of the following groups, or new "suits":

- Trump suit
- Suit agent has multiple cards in (not trumps)
- Suit in agent's highest singleton
- Suit in agent's second-highest singleton
- Suit in agent's third-highest singleton

This allows for a much more efficient description of the agent's hand, greatly improving the dimensionality of the inputs, and hence the learning rate of the agents. These five options are encoded in a binary format, for stability purpose, and hence three binary inputs are required to represent the suits. To represent the card's number, ten discrete values must be represented, hence requiring four binary inputs to represent the card's value. Thus a card in an agent's hand is represented by seven binary inputs, as depicted in Fig. 4.

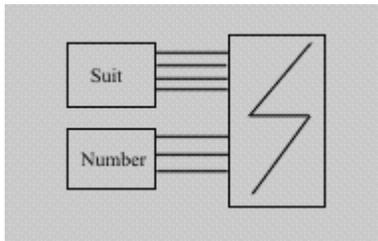

Fig. 4. Agent card input structure

Next must be considered the information required in order to make decisions two and three. For both of these decisions, the cards that have already been played, if any, are necessary to know in order to make an intelligent decision as to the correct next card to play. For the second decision, it is also plausible that knowledge of who has won a trick would be important. The most cards that can ever be played before a decision must be made is seven, and since the table after a card is played is used to evaluate and update the network, eight played cards are necessary to be represented. Once again, however, simply utilising the obvious encoding method is not necessarily the most efficient method. The actual values of the cards played are not necessarily important, only their values relative to the cards in the agent's hand. As such, the values can be represented as one of the following, with respect to the cards in the same suit in the agent's hand:

- Higher than the card/cards in the agent's hand
- Higher than the agent's second-highest card
- Higher than the agent's third-highest card
- Lower than any of the agent's cards
- Member of a void suit (number is immaterial)

Also, another suit is now relevant for representation of the played cards, namely a void suit – a suit in which the agent has no cards. Lastly, a number is necessary to handle the special case of the Ace of trumps, since its unique rules mean that strategies are possible to develop based on whether it has or has not been played. The now six suits available still only require three binary inputs to represent, and the six number groupings now reduce the value representations from four binary inputs to three binary inputs, once again reducing the dimensionality of the input system.

With all of these inputs specified, the agent now has available all of the information required to draw its own conclusions and create its own strategies, without human-imposed assumptions affecting its "thought" patterns.

*2) Network Architecture Design:* With the inputs now specified, the hidden and output layers need to be designed. For the output neurons, these need to represent the prediction P that the network is making [2]. A single hand has one of five possible outcomes, all of which need to be catered for. These possible outcomes are:

- The agent wins all three tricks, winning 3 chips.
- The agent wins two tricks, winning 2 chips.
- The agent wins one trick, winning 1 chip.
- The agent wins zero tricks, losing 3 chips.
- The agent elects to fold, winning no tricks, but losing no chips.

This can be seen as a set of options, namely [-3 0 1 2 3]. While it may seem tempting to output this as one continuous output, there are two compelling reasons for breaking these up into binary outputs. The first of these is in order to optimise stability, as elaborated upon in Section five. The second reason is that these are discrete events, and a continuous representation would cover the range of [-3 0], which does not in fact exist. The binary inputs then specified are:

- $P(O = 3)$
- $P(O = 2)$
- $P(O = 1)$
- $P(O = -3)$

With a low probability of all four catering to folding, winning and losing no chips. Consequently, the agent's predicted return is:
$$P = 3A + 2B + C - 3D \tag{1}$$

where
$$A = P(O = 3) \tag{2}$$

$$B = P(O = 2) \quad (3)$$
$$C = P(O = 1) \quad (4)$$
$$D = P(O = -3) \quad (5)$$

The internal structure of the neural network uses a standard sigmoidal activation function [7], which is suitable for stability issues and still allows for the freedom expected from a neural network. The sigmoidal activations function varies between zero and one, rather than the often-used one and minus one, in order to optimise for stability [7]. Since a high degree of freedom is required, a high number of hidden neurons are required, and thus fifty have been used. This number is iteratively achieved, trading off training speed versus performance. The output neurons are linear functions, since while they represent not binary effects, but rather a continuous probability of particular binary outcomes

*2) Agent Decision Making:* With its own predictor specified [2], the agent is now equipped to make decisions when playing. These decisions are made by predicting the return of the resultant situation arising from each legal choice it can make. An ε-greedy policy is then used to determine whether the agent will choose the most promising option, or whether it will explore the result of the less appealing result. In this way, the agent will be able to trade off exploration versus exploitation.

## IV. THE INTELLIGENT MODEL

With each agent implemented as described above, and interacting with each other as specified in Section III, we can now perform the desired task, namely that of utilising a multi-agent model to analyse the given game, and develop strategies that may "solve" the game given differing circumstances. Only once agents know how to play a certain hand can they then begin to outplay, and potentially bluff each other.

### A. Agent Learning Verification

In order for the model to have any validity, one must establish that the agents do indeed learn as they were designed to do. In order to verify the learning of the agents, a single intelligent agent was created, and placed at a table with three 'stupid' agents. These 'stupid' agents always stay in the game, and choose a random choice whenever called upon to make a decision. The results show quite conclusively that the intelligent agent soon learns to consistently outperform its opponents, as shown in Fig. 5.

The agents named Randy, Roderick and Ronald use random decision-making, while AIden has the TD(λ) AI system implemented [5]. The results have been averaged over 40 hands, in order to be more viewable, and to also allow for the random nature of cards being dealt. As can be seen, AIden is consistently performing better than its counterparts, and continues to learn the game as it plays.

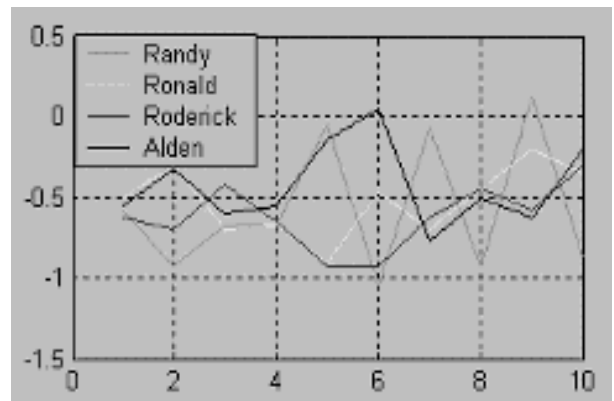

Fig. 5. Agent performance, averaged over 40 hands

*1) Cowardice:* In the learning phase of the abovementioned intelligent agent, an interesting and somewhat enlightening problem arises. When initially learning, the agent does not in fact continue to learn. Instead, the agent quickly determines that it is losing chips, and decides that it is better off not playing, and keeping its chips! This is illustrated in Fig. 6.

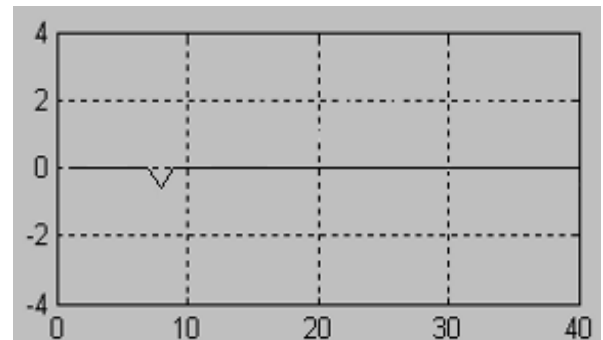

Fig. 6. Agent cowardice. Averaged over 5 hands

As can be seen, AIden [8] quickly decides that the risks are too great, and does not play in any hands initially. After forty hands, AIden decides to play a few hands, and when they go badly, gets scared off for good. This is a result of the penalising nature of the game, since bad play can easily mean one loses a full three chips, and since the surplus of lost chips is nor carried over in this simulation, a bad player loses chips regularly. While insightful, a cowardly agent is not of any particular use, and hence the agent must be given enough 'courage' to play, and hence learn the game. In order to do this, one option is to increase the value of ε for the ε-greedy policy, but this makes the agent far too much like a random player without any intelligence. A more successful, and sensible solution is to force the agent to play when it knows nothing, until such a stage as it seems prepared to play. This was done by forcing AIden [8] to play the first 200 hands it had ever seen, and thereafter leave AIden to his own devices [8], the result of which has been shown already in Fig. 5.

## B. Parameter Optimisation

A number of parameters need to be optimised, in order to optimise the learning of the agents. These parameters are the learning-rate α, the memory parameter λ and the exploration parameter ε. The multi-agent system provides a perfect environment for this testing, since four different parameter combinations can be tested competitively. By setting different agents to different combinations, and allowing them to play against each other for an extended period of time (number of hands), one can iteratively find the parameter combinations that achieve the best results, and are hence the optimum learning parameters [3]. Fig. 7 shows the results of one such test, illustrating a definite 'winner', whose parameters were then used for the rest of the multi-agent modeling. It is also worth noting that as soon as the dominant agent begins to lose, it adapts its play to remain competitive with its less effective opponents. This is evidenced at points 10 and 30 on the graph (games number 300 and 900, since the graph is averaged over 30 hands) where one can see the dominant agent begin to lose, and then begins to perform well once again.

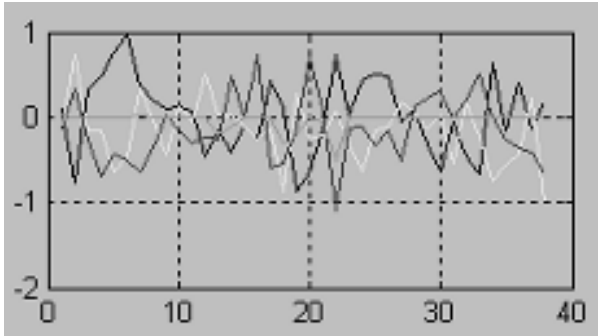

Fig. 7. Competitive agent parameter optimisation. Averaged over 30 hands

Surprisingly enough, the parameters that yielded the most competitive results were α = 0.1; λ = 0.1 and ε = 0.01. while the ε value is not particularly surprising, the relatively low α and λ values are not exactly intuitive. What they amount to is a degree of temperance, since higher values would mean learning a large amount from any given hand, effectively over-reacting when they may have played well, and simply have fallen afoul of bad luck.

## C. MAS Learning Patterns

With all of the agents learning in the same manner, it is noteworthy that the overall rewards they obtain are far better than those obtained by the random agents, and even by the intelligent agent that was playing against the random agents [3]. A sample of these results is depicted in Fig. 8.

R1 to R3 are the Random agents, while AI1 is the intelligent agent playing against the random agents. AI2 to AI 5 depict intelligent agents playing against each other. As can be seen, the agents learn far better when playing against intelligent opponents, an attribute that is in fact mirrored in human competitive learning.

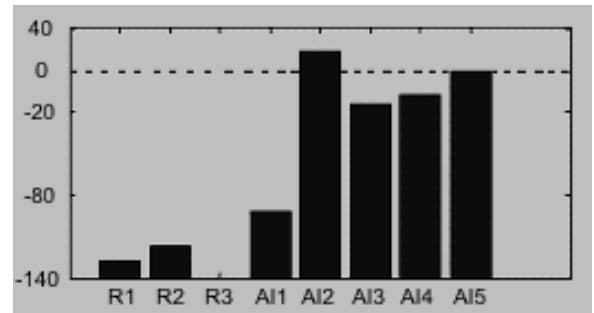

Fig. 8. Comparative returns over 200 hands

The agents with better experience tend to fold bad hands, and hence lose far fewer chips than the intelligent agent playing against unpredictable opponents.

## D. Agent Adaptation

In order to ascertain whether the agents in fact adapt to each other or not, the agents were given pre-dealt hands, and required to play them against each other repeatedly. The results one such experiment, illustrated in Fig. 9, shows how an agent learns from its own mistake, and once certain of it changes its play, adapting to better gain a better return from the hand. The mistakes it sees are its low returns, returns of -3 to be precise. At one point, the winning player obviously decides to explore, giving some false hope to the losing agent, but then quickly continues to exploit his advantage. Eventually, at game #25, the losing agent gives up, adapting his play to suit the losing situation in which he finds himself. Fig. 9 illustrates the progression of the agents and the adaptation described.

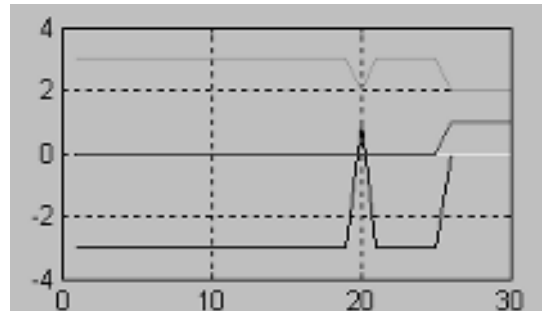

Fig. 9. Adaptive agent behaviour

## E. Strategy Analysis

The agents have been shown to successfully learn to play the game, and to adapt to each other's play in order to maximise their own rewards. These agents form the pillars of the multi-agent model, which can now be used to analyse, and attempt to 'solve' the game. Since the game has a nontrivial degree of complexity, situations within the game are to be solved, considering each situation a sub-game of the overall game. The first and most obvious type of analysis is a static analysis, in which all of the hands are pre-dealt. This system can be said to have stabilised when the results and the playout become constant, with all agents content to play the hand out in the same manner, each deciding nothing better can be achieved. This is akin to Game Theory's "static equilibrium" [4], as is evidenced in Fig. 10.

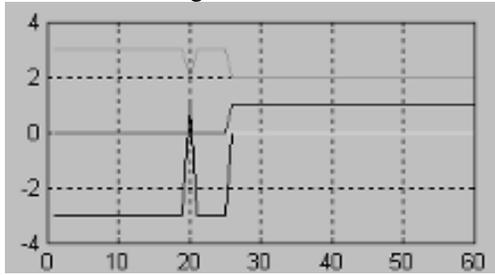

Fig. 10. Stable, solved hand

*F. Bluffing*

A bluff is an action, usually in the context of a card game that misrepresents one's cards with the intent of causing one's opponents to drop theirs. There are two opposing schools of thought regarding bluffing. One school claims that bluffing is purely psychological, while the other maintains that a bluff is a purely statistical act, and therefore no less sensible than any other strategy. Astoundingly enough, the intelligent agents do in fact learn to bluff! A classic example is illustrated in Fig. 11, which depicts a hand in which bluffing was evidenced

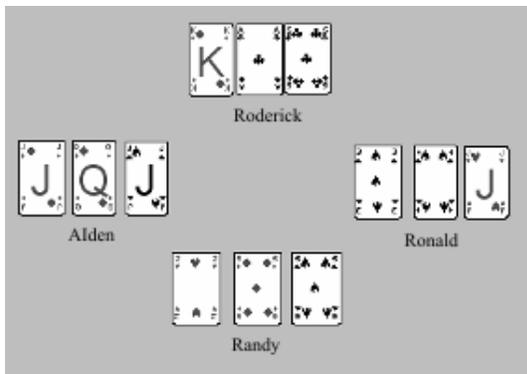

Fig. 11. Agent bluffing

In the above hand, Randy is the first caller, and diamonds have been declared trumps. Randy's hand is not particularly impressive, having only one low trump, and two low supporting cards. Still, he has the lead, and a trump could become a trick, although his risks are high for minimal reward. Nonetheless, Randy chooses to play this hand. Ronald, having nothing to speak of, unsurprisingly folds. Roderick, on the other hand, has a very good hand. One high trump, and an outside ace. However, with one still to call, and Randy already representing a strong hand by playing, Roderick chooses to fold. AIden, whose hand is very strong with two high trumps and an outside jack, plays the hand. When the hand is played repeatedly, Randy eventually chooses not to play, since he loses all three to AIden. Instantly, Roderick chooses to play the hand, indicating that the bluff was successful, that it chased a player out of the hand! Depending on which of the schools of thought regarding bluffing one follows this astonishing result leads us to one of two possible conclusions. If, like the author, one hold that bluffing is simply playing the odds, making the odds for one's opponent unfavourable by representing a strong hand, then this result shows that the agents learn each other's patterns well enough to factor their opponent's strategies into the game evaluation, something game theory does a very poor job of. Should one follow the theory that bluffing is purely psychological, then the only conclusion that can be reached from this result is that the agents have in fact developed their own 'psyches', their own personalities which can then be exploited. Regardless of which option the reader holds to, the fact remains that agents have been shown to learn, on their own and without external prompting, to bluff!

V. CONCLUSIONS

While the exact nature of bluffing is still unknown, it has been shown that a system involving agents capable of learning adaptively not only from the game being played, but also from their opponents, is in fact able to learn to predict its opponent's reactions. This knowledge in turn changes the statistical nature of a game being played, allowing agents to learn to bluff, based purely on rational reasoning, lending strong support to the theory that bluffing is simply *playing the odds*, and not an illogical, psychologically based action. The use of the Reinforcement learning paradigm [1, 6], along with the TD($\lambda$) algorithm [5] for adaptively training neural networks, ahs been shown to meet all of the requirements to produce such agents. Lastly, the design of the agent "view", has been seen to be the most important facet of creating bluffing agents, since their view of the game as inclusive of the other players allows for the incorporation of those players into it's estimation of the game's outcome. With all of these steps adhered to, artificially intelligent agents can learn to bluff!

REFERENCES

[1] R.S. Sutton, and A.G. Barto, *Reinforcement learning: An Introduction*, MIT press, 1998.
[2] R.S. Sutton, "Learning to predict by the methods of temporal differences". *Machine Learning*, vol 3, pp. 9-44, 1988.
[3] E. Hurwitz, and T. Marwala, Optimising Reinforcement Learning for Neural Networks, *In Proceedings of the 6th Annual European on Intelligent Games and Simulation*, Leicester, UK, pp. 13-18, 2005,.ISBN: 90-77381-23-6.
[4] A.H. Morehead, G. Mott-Smith, and P.D. Morehead, *Hoyle's Rules of Games, Third revised and updated edition*. Signet book. 2001.
[5] R.S. Sutton, *Implementation details of the TD($\lambda$) procedure for the case of vector predictions and Backpropogation*. GTE laboratories technical note TN87-509.1, 1989.
[6] R.S. Sutton, Frequenlty asked questions about reinforcement learning, http://www.cs.ualberta.ca/~sutton/RL-FAQ.html: last viewed 24/08/2005,
[7] O. Smart, *Line Search and Descents Methods*. http://www.biochemistry.bham.ac.uk/osmart/msc/lect/lect2a.html 02/02/1996. Last Viewed 20/04/2006.
[8] J. Wesley Hines, *Matlab supplement to Fuzzy and Neural Approaches in Engineering*. John Wiley & Sons Inc. New York. 1997.
[9] Miller, Sutton and Werbos. *Neural Networks for Control.* MIT Press. Cambridge, Massecheussettes. 1990.